\pdfoutput=1

\documentclass[11pt]{article}
\usepackage{amsmath}

\usepackage{acl}
\usepackage{pip}

\usepackage{times}
\usepackage{latexsym}
\usepackage{linguex}
\usepackage{graphicx}
\usepackage{booktabs}
\usepackage{multirow, makecell, colortbl, hhline, adjustbox}
\usepackage{caption}
\usepackage{subcaption}
\usepackage{cellspace}
\usepackage{xcolor}
\usepackage[inkscapelatex=false]{svg}

\newcommand{\ethz}{1}
\newcommand{\google}{2}

\usepackage[T1]{fontenc}

\usepackage[utf8]{inputenc}

\usepackage{microtype}

\usepackage{inconsolata}

\title{In-Context Probing: Toward Building Robust Classifiers via Probing Large Language Models}

\author{Afra Amini$^{\ethz, \google, }$\thanks{\hspace{0.5em} Work done during internship at Google DeepMind.} \qquad Massimiliano Ciaramita$^{\google}$\\
  \hspace{-1.8cm}$^{\ethz}$ETH Zurich \qquad \quad $^{\google}$Google DeepMind\\
  }

\begin{document}
\maketitle

\begin{abstract}
Large language models are able to learn new tasks \emph{in context}, where they are provided with instructions and a few annotated examples. 
However, the effectiveness of in-context learning is dependent on the provided context, and the performance on a downstream task can vary considerably, depending on the instruction. Importantly, such dependency on the context can surface in unpredictable ways, e.g., a seemingly more informative instruction might lead to a worse performance. In this paper, we propose an alternative approach, which we term \defn{In-Context Probing} (ICP). Similar to in-context learning, we contextualize the representation of the input with an instruction, but instead of decoding the output prediction, we probe the contextualized representation to predict the label. Through a series of experiments on a diverse set of classification tasks, we show that in-context probing is significantly more robust to changes in instructions. We further show that ICP performs competitive or superior to finetuning and can be particularly helpful to build classifiers on top of smaller models, with less than a hundred training examples. 
\end{abstract}

\section{Introduction}
Language models have become an essential tool in 
dealing with various tasks in the natural language processing (NLP) domain \citep[][\emph{inter alia}]{howard-ruder-2018-universal, devlin-etal-2019-bert, NEURIPS2020_1457c0d6}. Notably, such models are extremely sample efficient, and can be employed to solve downstream tasks with a small set of labeled data. The conventional pipeline to build classifiers is to finetune language models to solve the task at hand.

\begin{figure}[t]
    \centering
    \includegraphics[width=\columnwidth]{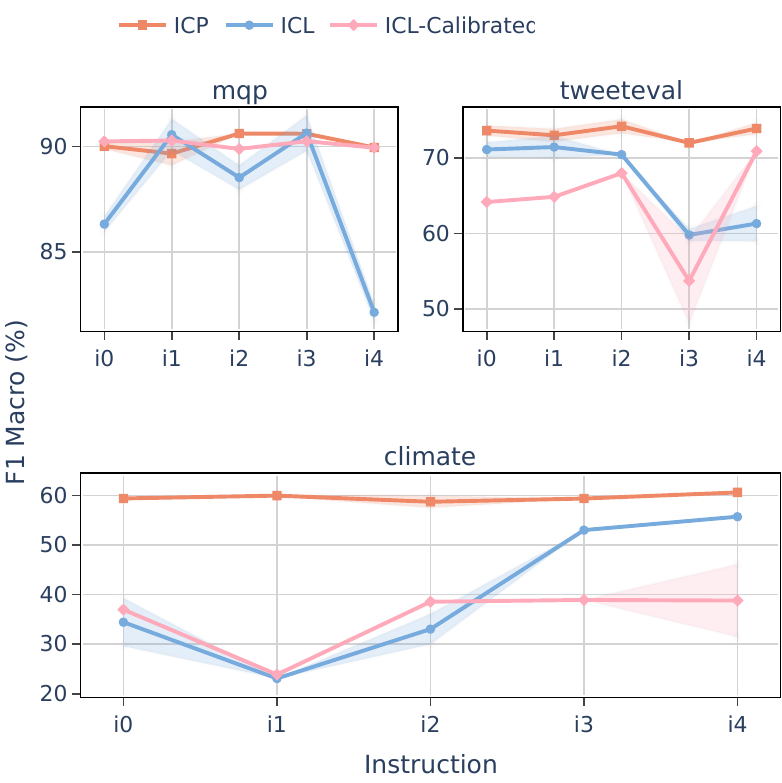}
    \caption{Comparing the robustness of in-context learning and in-context probing to the change in instructions. The x-axis shows different instructions formulations, and the y-axis shows F1 Macro score. Traces visualize the mean and standard deviation of the performance with $5$ different seeds for sampling the training examples (or demonstrations for ICL). ICP performance is significantly more robust to variations in the instructions compared to ICL and Calibrated ICL in \climate{} and \tweeteval{} tasks.}
    \label{fig:q1}
\end{figure}
\begin{figure*}[t]
    \centering
    \includegraphics[width=0.85\linewidth]{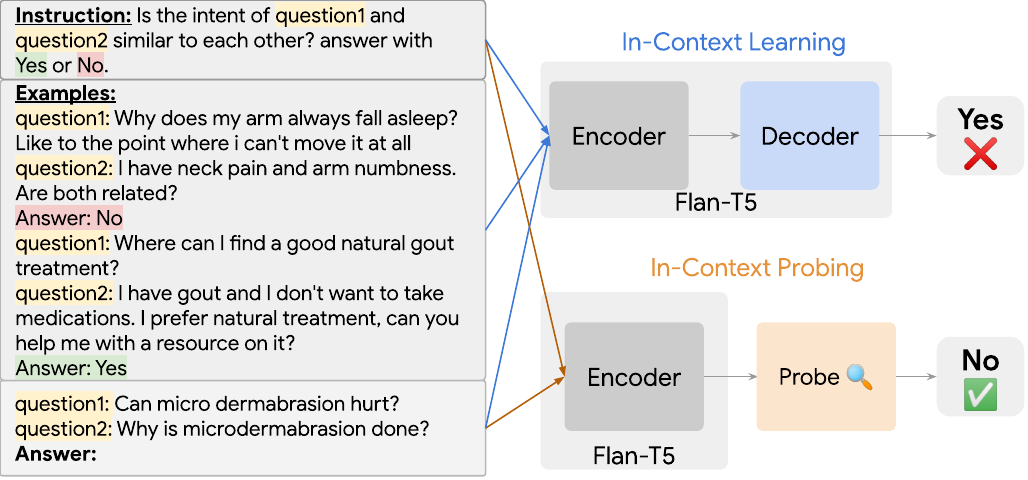} 
    \caption{The figure demonstrates and compares in-context probing and in-context learning workflows. The example is from the medical question pairs dataset \citep{mqp}. In-context probing receives contextualized representations of instruction and the input, and directly predicts the classification label.}
    \label{fig:example}
\end{figure*}

However, as the language models grow in number of parameters, finetuning them becomes more computationally expensive. Moreover, finetuning changes a model's parameters through gradient updates. Therefore, for each downstream task, a new model must be trained and stored. Fortunately, it has been shown that as models grow in size, they can learn \emph{``in context''} \citep{radford2019language, NEURIPS2020_1457c0d6}. \defn{In-Context Learning} (ICL) refers to prompting a language model with a few demonstrative examples. Such prompting 
bundles up a small set of input-label pairs into: a set of instructions, a few solved examples, and a single unlabeled example. The model is then asked to predict the label for that example.

Notably, since ICL does \emph{not} require any weight updates on the language model, a single model can be used to perform a variety of tasks, as long as they can be specified in natural language. While effective, there are some caveats to ICL that we elaborate next. 

First, zero-shot, or few-shot, performance of pretrained models on downstream tasks depends to a large extent on the way in which the prompt is formulated \citep{pmlr-v139-zhao21c,liu-prompt-survey-2023}. 

Second, ICL may not always benefit from the demonstrative examples the way that it is expected to \citep{webson-pavlick-2022-prompt}, e.g., flipping the labels in demonstrations rarely hurts the performance \citep{min-etal-2022-rethinking}. Furthermore, the number of demonstrative examples that ICL \emph{can} benefit from is bounded by the language model constraint on the maximum length of input sequences.

Lastly, decoding the predicted label is neither accurate nor efficient. The decoder might be miscalibrated \citep{pmlr-v139-zhao21c} or suffer from surface form competition \citep{holtzman-etal-2021-surface}. Moreover, autoregressive decoding, with an attention-based decoder takes $\bigo(N^2)$ time for generating a sequence of length $N$, which can be inefficient when decoding long sequences. 

To address these limitations, we investigate an alternative workflow to ICL for building accurate and robust classifiers. We postulate that a primary cause of the aforementioned limitations, such as the sensitivity to the exact phrasing of the instruction and miscalibration, is that the model is forced to verbalize the prediction label. Therefore, we suggest to bypass the decoding step and directly \emph{probe} the extracted representations from pretrained models. We hypothesize that given reasonable instructions, the information that is needed to \emph{reliably} perform the downstream task is encoded by the model in contextualized representations of the tokens. Notably, we contextualize the input by providing instructions. We therefore name our approach in-context probing (ICP). We demonstrate our proposed workflow and contrast it with ICL through an example in \Cref{fig:example}.


Through an extensive set of experiments on a diverse set of sentence classification tasks and different model sizes, we aim to answer the following research questions:

\begin{itemize}
\itemsep0em
    \item \textbf{Q1:} Is in-context probing more robust with respect to variations in instructions compared to ICL? (Section \cref{sec:q1})
    \item \textbf{Q2:} Can in-context probing perform classification tasks as accurately as ICL or even finetuning? (Sections \cref{sec:q2}, \cref{sec:finetune})
    \item \textbf{Q3:} Is in-context probing sample efficient? (Section \cref{sec:sample-eff})
\end{itemize}

We find that in-context probing is significantly less sensitive to subtle changes in instructions compared to in-context learning. We further compare in-context probing with in-context learning on different sizes of \flant{} models \citep{chung2022scaling} in \cref{sec:q2}. Our results suggest that for larger models, ICP is on par or better than both ICL and calibrated ICL with less variance to instruction changes. For smaller model sizes, ICP significantly outperforms ICL. Furthermore, we empirically show that ICP is sample efficient, as it can generate competitive results to ICL (with significantly less variance to instructions) after training only on $40$ annotated examples. Finally, comparing in-context probing with finetuning suggests that probing classifiers can be as accurate and robust as finetuned models, while using $4$ to $6$ orders of magnitude less trainable parameters. 

\section{Finetuning}
The typical workflow for adapting a language model to a new downstream task is finetuning. Specifically, a model is first pretrained on large text corpora in a self supervised manner. It is then finetuned on a downstream task by initializing with the pretrained weights, and adding a task-specific layer to be trained from scratch \cite{devlin-etal-2019-bert}.

\subsection{Parameter-Efficient Finetuning (PEFT)}
As language models grow in size, finetuning them for any given downstream task becomes increasingly expensive, if not infeasible. Hence, a growing body of research has focused on how to enable models to perform downstream tasks with reduced weight updates. Early works on PEFT, includes developing complementary network components such as \defn{adapters}; trainable feed-forward networks inserted between the layers of pretrained models \citep{NIPS2017_e7b24b11, pmlr-v97-houlsby19a, bapna-firat-2019-simple}. Following works aim to reduce the number of trainable parameters \citep{mahabadi2021compacter, hu2022lora}.
One can also view \defn{prompt tuning} \citep{lester-etal-2021-power} and \defn{prefix tuning} \citep{li-liang-2021-prefix} through the lens of PEFT \citep{he2022towards}, where learnable parameters are added to the model's inputs or activations, and are trained on a downstream task. 
While PEFT methods reduce the memory and computation cost, compared to finetuning, they still do require training $0.1$-$15\%$ of the models' parameters and assume access to models' weights. 
\subsection{Prompt-Based Finetuning}
An alternative approach to the finetuning workflow is to adapt the pretrained language model directly by autoregressively decoding the target output. For instance, in a classification task, one could prompt the language model by asking for the classification outcome, and tune the model's parameters with the goal of decoding the gold label. Notably, it has been shown that prompt-based finetuning of text-to-text transformers, e.g., \citep{t5}, leads to competitive results in classification benchmarks such as SuperGlue \citep{superglue}.

Moreover, prompt-based finetuning is extremely effective in low- to medium-resourced data regimes \citep{gao-etal-2021-making, le-scao-rush-2021-many} and offers a decent out-of-distribution performance \citep{mosbach2023fewshot}.

\section{In-Context Learning} 
While effective, finetuning and storing large language models with billions of parameters is often impractical. In-context learning offers an alternative approach to learn new downstream tasks without any weight updates. The general strategy is to prompt the language model not only with the input, but also with the instructions required to solve the task and a few demonstrative examples of input-target pairs, all written in natural language. 

Consider the example in \Cref{fig:example}. The prompt consists of an instruction, two demonstrations, and an input. The language model is expected to learn the task from the provided context. The output label is obtained by autoregressively decoding the model prediction.

\paragraph{Instruction Finetuning.} To boost the performance of ICL, especially in zero-shot settings, finetuning the models with instruction is helpful. The goal of the instruction finetuning step is to teach the model to benefit the most from in-context learning. Importantly and as opposed to pattern-based finetuning, instruction finetuning is done only \emph{once} over a diverse set of tasks with instructions, with the aim to teach the model how to follow instructions and transfer this learning to unseen tasks. \citet{wei2022finetuned} show that after finetuning models with instructions, the zero-shot performance on unseen tasks improves significantly. \citet{chung2022scaling} further scale this approach and introduce and release $\flant$ checkpoints that achieve strong few-shot performance compared to larger models. 
\subsection{Contextual Calibration} 
A major limitation of ICL is that its performance can depend crucially on the way instructions are formulated, the order of the demonstration examples, and the unbalanced distribution of labels. To reduce this dependency, \citet{pmlr-v139-zhao21c} proposed calibrated ICL. 

Calibration works as follows: First, a \defn{content-free prompt} is created. This prompt includes the same instruction and demonstration examples as the original prompt, but the input data is removed.\footnote{One can replace the input data with an empty string, ``\texttt{N/A}'', or gibberish tokens. In our experiment we remove the input data by replacing it with an empty string.} The main idea is that the model's bias toward predicting certain labels is revealed by its output for the content-free prompt.

Next, the predicted probabilities of labels are calibrated by the predicted probabilities for the content-free prompt. Concretely, suppose the model predicts $p(c {\mid} \prompt)$, which shows the probability of the input being classified with label $c$. To calibrate this probability, we divide it by the probability of predicting the same label $c$ for the content-free prompt: $\frac{p(c \mid \prompt)}{p(c \mid \prompt_{null})}$, where $\prompt_{null}$ is obtained by removing the input data from the prompt $\prompt$. The probabilities are then normalized via Softmax and the class with the maximum probability is predicted as the label. 

\paragraph{Limitations.} Although ICL is an efficient alternative to finetuning, with competitive results, it has its own limitations. As mentioned earlier, and even after calibration, ICL is notoriously unstable \citep{lu-etal-2022-fantastically}, minor changes in the prompt can drastically affect the prediction results. Furthermore, as pointed out by \citep{min-etal-2022-rethinking}, ICL might be limited to learning the domain of input and the label, and not the exact mapping between the input and the label. Similarly, \citet{webson-pavlick-2022-prompt} questions the degree to which the effectiveness of ICL is derived from models understanding task instructions in ways analogous to humans.
\begin{figure*}[t]
     \centering
     \includegraphics[width=2\columnwidth]{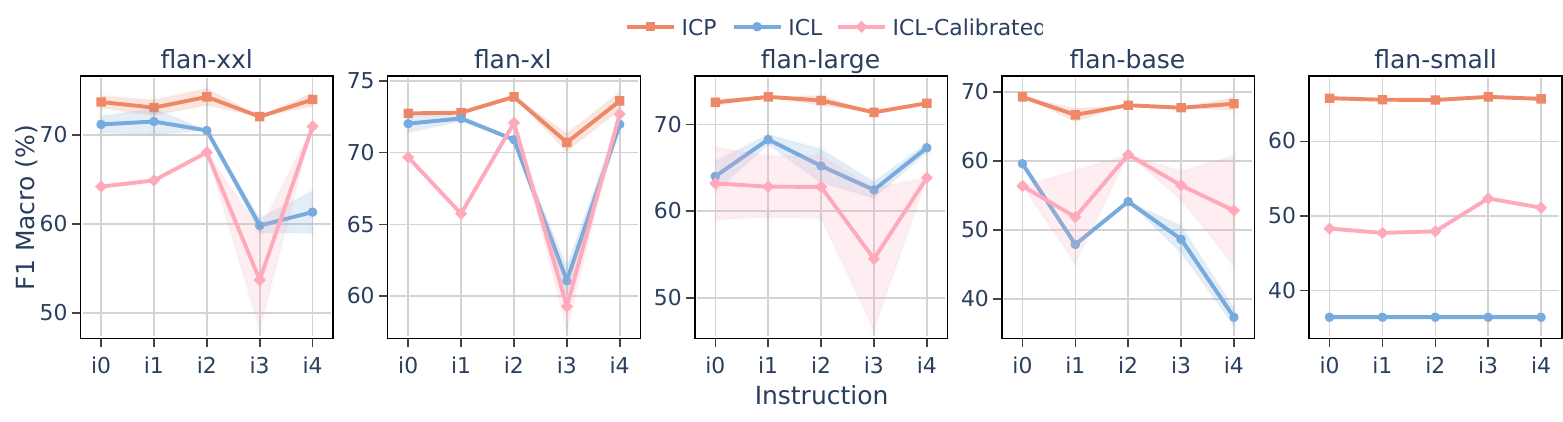}
     \caption{Comparing the robustness of the three approaches on \tweeteval{} task across different model sizes. We observe that the performance of ICP is close or better than ICL, while being significantly more robust to the change of instructions.}
     \label{fig:robust-models}
 \end{figure*}
\paragraph{Our proposal} departs from these two lines of work, i.e., finetuning and in-context learning. We provide a more efficient approach to finetuning which, as opposed to finetuning methods, does \emph{not} require any knowledge about a models' architecture, can be trained with as few as a hundred examples, and only requires access to the last layer output of a model. 

The main observation is that, similar to in-context learning, we can benefit from instructions to extract useful contextualized representations from large language models to perform a downstream task. However, and different to ICL, we make use of direct training signals when training our light-weight probes. We empirically show that in-context probing significantly reduces the performance dependency to the formulation of instructions. We elaborate on our probing approach in the next section.

\section{Probing}
Probes, or diagnostic classifiers, are light-weight classifiers trained to predict a property of interest from representations extracted from a model \citep{alain2016understanding, hupkes2018visualisation}. Originally, probes were used as a diagnostic tool, i.e., to assess whether or not information about a property is encoded in the representations. In this section, we explain how to re-purpose probes from diagnostic tools to accurate and robust classifiers. 

Given an input prompt sequence $\prompt = [x_1, x_2, \dots, x_N]$, which consists of an instruction and an input, a transformer-based encoder generates a sequence of representations $\reps = [\rep_1, \rep_2, \dots, \rep_N] \in \mathbb{R}^{N \times d}$, where $d$ is the dimensionality of the encoder model.\footnote{In this work, we probe the representations extracted from the last layer of the encoder. However, one can probe decoder-only models in a similar manner.} A probe is a function $\probe$ which receives input representations $\reps$ and predicts the index of the label $m \in \{1, 2, \dots, M\}$, where $M$ is the number of classification labels. Next, we introduce two architectures that we evaluate for building probing classifiers.
\paragraph{Logistic Regression Probe.} A logistic regression probe receives representations $\reps$ for an input sequence $\prompt$, and predicts the classification label as
\begin{equation} \label{eq:probe}
    \probe(\reps) = \underset{m}{\mathrm{argmax}}\, \{\weights_m^T (\sum_{i=1}^N \rep_i) + \bias_m\}
\end{equation}
Where $\bm{\theta} = (\weights \in \mathbb{R}^{d \times M}, \bias \in \mathbb{R}^M)$ are learned parameters of the probing model. 
\paragraph{Attentional Probe.} Our attentional probe is a single attention layer (with one attention head). It recieves the sequence of contextualized representations and predicts the classification label as
\begin{equation} 
     \probe(\reps) = \underset{m}{\mathrm{argmax}}\, \{\weights_m^T (\sum_i \alpha_i \rep_i) + \bias_m\}
 \end{equation}
Where $\alpha_i$s are attention weights: 

\begin{equation}
    \alpha_{i} = (\key \rep_i)^T (\query \rep_0)
\end{equation}

The learned parameters are $\bm{\theta} = (\key, \query, \weights, \bias)$, where $\key$ and $\query$ are key and query matrices. $\rep_0$ is the fixed contextualized representation of the \textit{``instruction''} token.\footnote{Instead of using a fixed query vector, we could also learn one. However, we empirically find that even without adding extra parameters for the query vector to the probing model, we achieve competitive results to ICL.} To train both probing models, we minimize the cross entropy loss.

\section{Experimental Setup}
\paragraph{Dataset.} We experiment with 3 sentence classification tasks: (i) paraphrase detection \citep[\mqp{};][]{mqp} where the goal is to predict whether two medical questions are semantically equivalent or not, (ii) natural language inference \citep[\climate{};][]{climate} where the goal is to predict whether the evidence entails the statement about climate change, refutes it, or is neutral to it, (iii) hate speech detection \citep[\tweeteval{};][]{tweeteval}, where the goal is to detect whether hate speech is expressed in a given tweet or not. See \cref{tab:dataset} for datasets statistics.

The datasets are chosen based on multiple reasons. First, none of them are included in the instruction finetuning of the \flant{}  models. Second, all of them are quite low-resourced, with $\approx 10k$ training examples. Lastly, these datasets cover different types of classification tasks.

\paragraph{Model.} We experiment with different model sizes of the \flant{} family, namely: \flans(80M), \flanbase(250M), \flanl(780M), \flanxl(3B), and \flanxxl(11B).  

\paragraph{Probing Model.} For training probes and finetuning models, we use $30\%$ of training examples for validation. We stop the training (or finetuning) process early, if the validation performance does not improve after $5$ consequent epochs, and report the test-set performance of the model with the best performance on the validation set. See \cref{sec:details} for more details. Throughout the paper, when the architecture of the probe is not explicitly mentioned, the least complex probe, i.e., the logistic regression probe is used. We compare the performance of the two probes in \cref{sec:arch}. Following prior works \citep{min-etal-2022-rethinking}, we report macro F1 score for all the tasks.\looseness=-1

\section{Results}
In the following subsections we discuss the experiments and our findings. In \cref{sec:q1} we evaluate the robustness of in-context probing. In \cref{sec:q2} and \cref{sec:sample-eff} we assess the performance and sample efficiency of probes, respectively. In \cref{sec:finetune} we compare in-context probing with finetuning. Finally, in \cref{sec:arch} we ablate the probing model's architecture.

\subsection{Robustness} \label{sec:q1}
To test the robustness of ICP and ICL with the change of instructions, we manually write 5 instructions per task (see \Cref{tab:inst}). Notably, the instructions are \emph{not} designed to be adversarial, but only to explain the task in different terms. 
\begin{figure*}[t]
    \centering
    \includegraphics[width=2\columnwidth]{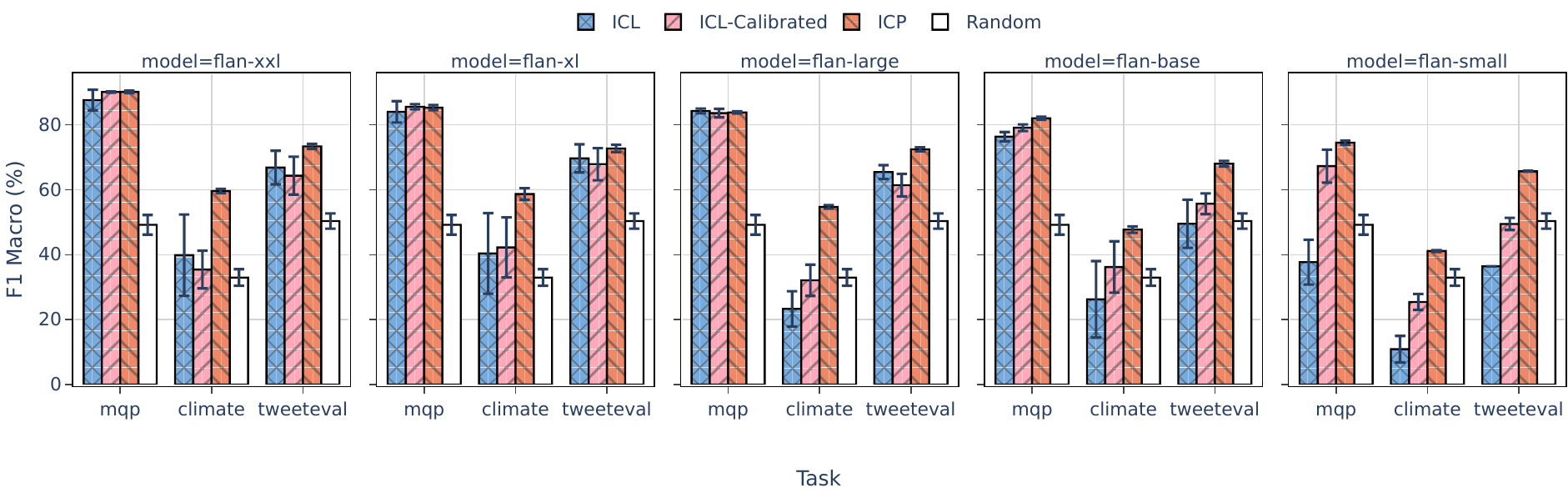}
    \caption{Comparing the performance of ICL, Calibrated ICL, and In-Context Probing (ICP). Error bars show the variance when using different instructions. ICP consistently performs better or on par with ICL and with lower variance across different model sizes and tasks. For smaller-sized models, ICP significantly outperforms both ICL and Calibrated ICL in \climate{} and \tweeteval{} tasks.}
    \label{fig:q2}
\end{figure*}
We sample examples from the training set with $5$ random seeds, and compare ICL with ICP on the $3$ classification tasks. To implement ICL, we use these training examples as demonstrations in the prompt, and decode the classification label.

For ICP, we do \emph{not} use any demonstrations, and only provide the model with instructions and the input. To reduce the clutter in the plots, we only visualize the performance of the methods using the number of training examples that lead to the \emph{best} performance. That is an average of $3$ demonstrations for the in-context learning method, $2$ for calibrated ICL, and $170$ training examples for the ICP approach (we will discuss the sample efficiency in \Cref{sec:sample-eff}).\looseness=-1

First, we compare the performance of the three methods (ICL, Calibrated ICL, and ICP) on the best-performing model in the mix, i.e., \flanxxl{} model. We visualize the variation of the performance with respect to the instructions in \cref{fig:q1}. In this figure, traces show the mean performance across $5$ different samples of training data (or demonstration examples) and shades demonstrate the standard deviation of performance with respect to the selected samples. In general, we see that probes' performance is significantly more robust to the change of the instructions. Interestingly, ICL performance decreases to a close to chance accuracy with minor changes in the instruction on \climate{} task. For example, one of the instructions that leads to a low accuracy is: \texttt{Based on the evidence, can we conclude that the claim is definitely supported? Answer with only one of the following options: yes, supported | no, refuted | not enough info}, which is not substantially different to the instruction that leads to the best result: \texttt{Given the evidence, is the claim definitely supported, refuted, or there is not enough info? Answer with only one of the following options: supports | refutes | not enough info}. 
\begin{figure*}[h]
    \centering
    \includegraphics[width=2\columnwidth]{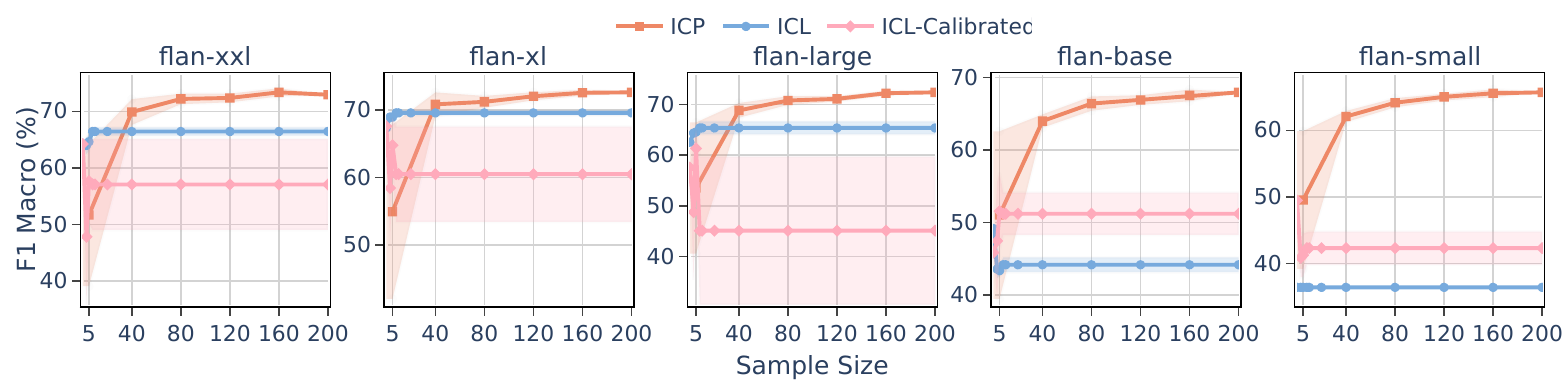}
    \caption{Comparing the performance of in-context learning and in-context probing with respect to number of training examples. Traces show the mean and standard deviation of the performance with 5 different seeds for the \tweeteval{} task. In all model sizes, training probes on only $40$ examples performs on par or better than ICL and calibrated ICL.}
    \label{fig:q3}
\end{figure*}

We observe that calibration is only helpful in one task, i.e., \mqp{}, where it reduces the standard deviation of performance with respect to instruction change. However, in \climate{} and \tweeteval{}, calibration is not helpful and the performance of ICL changes significantly with the change of instructions.

Next, we experiment with the effect of model size on the robustness of different methods in \cref{fig:robust-models}. In this plot we focus on the \tweeteval{} task.\footnote{Similar trends can be observed in other tasks and the plots for all tasks and model sizes can be found in \cref{fig:robust-all}.} Generally, we observe a similar trend across all model sizes: ICP performs consistently well, irrespective of the choice of instruction, but the performance of ICL and calibrated ICL fluctuates quite significantly. The exception to this trend is the smallest model (\flans{}), where we don't see a high standard deviation in the performance, when employing ICL and calibrated ICL methods. However, this is possibly because ICL consistently performs poorly compared to ICP. We will expand on this result in the next section by offering a broader view on the performance of the methods across models and tasks. 

\subsection{Overall Performance} \label{sec:q2}
We now look more holistically at the performance of in-context probing compared to in-context learning, across different tasks and model sizes. For each model size and task, we look at the performance of the $3$ methods using $5$ different seeds of sampling the training examples. We further show the standard deviation of the performance when varying the instructions. Same as the previous experiment, for all of the methods we pick the number of training examples to be the one that results in the highest performance. As results in \cref{fig:q2} suggest, for larger models, in-context probing achieves competitive results to ICL with significantly less variance. 

Importantly, for smaller models, the performance of the in-context learning approach is usually close to random predictions. in-context probing, however, is performing significantly better than random, and is even on par with the ICL scores of \emph{larger} models. This highlights the usefulness of in-context probing when resources are limited and only smaller models can be employed.
\begin{table}[t]
    \centering
    \adjustbox{width=\columnwidth}{%
    \begin{tabular}{@{}lp{2cm}ccc@{}}\toprule
    & \# parameters & \mqp{} & \tweeteval{} & \climate{} \\
    \rowcolor{myinst} \multicolumn{5}{c}{\flanxxl{}} \\
    FT & $11$B & $89.12\stdtiny{\mathbf{0.53}}$   & $71.0\stdtiny{3.14}$  & $54.60\stdtiny{3.67}$  \\
    ICP & $10$K & $\mathbf{89.95}\stdtiny{0.56}$  & $\mathbf{72.95}\stdtiny{\mathbf{0.59}}$  & $\mathbf{59.34}\stdtiny{\mathbf{1.21}}$ \\ 
    \rowcolor{myinst} \multicolumn{5}{c}{\flanxl{}} \\
    FT & $3$B & $\mathbf{87.89}\stdtiny{\mathbf{0.38}}$  &  $69.47\stdtiny{2.52}$ &  $50.91\stdtiny{\mathbf{1.49}}$  \\
    ICP & $6$K & $85.32\stdtiny{0.78}$  & $\mathbf{72.66}\stdtiny{\mathbf{1.1}}$ &  $\mathbf{57.45}\stdtiny{2.72}$\\ 
    \rowcolor{myinst} \multicolumn{5}{c}{\flanl{}} \\
    FT & $780$M & $\mathbf{86.23}\stdtiny{0.85}$  & $69.26\stdtiny{1.2}$   & $51.37\stdtiny{5.1}$ \\
    ICP & $3$K & $83.81\stdtiny{\mathbf{0.33}}$  & $\mathbf{72.42}\stdtiny{0.58}$ &  $\mathbf{53.75}\stdtiny{\mathbf{1.50}}$ \\ 
    \rowcolor{myinst} \multicolumn{5}{c}{\flanbase{}} \\
    FT & $250$M & $79.57\stdtiny{1.72}$ & $64.36\stdtiny{\mathbf{0.49}}$ &  $45.58\stdtiny{5.17}$  \\
    ICP & $2$K & $\mathbf{82.08}\stdtiny{\mathbf{0.59}}$ &  $\mathbf{67.92}\stdtiny{0.93}$ &  $\mathbf{47.71}\stdtiny{\mathbf{0.98}}$ \\ 
    \rowcolor{myinst} \multicolumn{5}{c}{\flans{}} \\
    FT & $80$M & $71.73\stdtiny{0.80}$ & $55.35\stdtiny{4.87}$ &  $34.98\stdtiny{1.63}$ \\
    ICP & $1$K & $\mathbf{74.44}\stdtiny{\mathbf{0.77}}$ &  $\mathbf{65.73}\stdtiny{\mathbf{0.17}}$ & $\mathbf{40.92}\stdtiny{\mathbf{0.47}}$ \\ \bottomrule
    \end{tabular}}
    \caption{Comparing the F1 Macro of in-context probing (ICP) to finetuning (FT) with $200$ training examples. The standard deviation (std) shows the performance change with respect to change of instructions used for finetuning or training the probe. While ICP has significantly less trainable parameters, it performs competitive (or superior) to finetuning, with comparable standard deviation across different instructions.}
    \label{tab:ft}
\end{table}

\begin{figure}[h]
    \centering
    \includegraphics[width=\columnwidth]{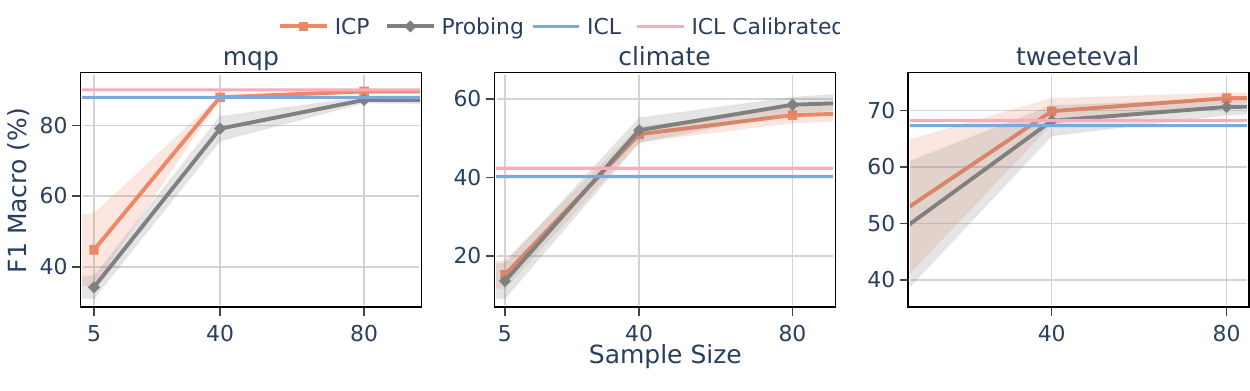}
    \caption{Comparing the sample efficiency of ICP to probing \flanxxl{} model. While both are generally efficient, ICP is significantly more sample efficient in \mqp{} task.}
    \label{fig:remove-inst}
\end{figure}

\subsection{Sample Efficiency} \label{sec:sample-eff}
Part of the effectiveness of in-context probing can be attributed to using more training signals compared to ICL. In this experiment, we look into probes' sample efficiency. Specifically, we define the sample efficiency of ICP as the number of samples needed for in-context probing to achieve a performance on par, or better than ICL.  We vary the sample size between $0$ to $200$.\footnote{In-context learning method, however, can only roughly benefit from less than $10$ examples, and longer inputs are truncated due to the limited context window of the models.} The number of used training samples is limited to $200$ for two reasons: first to evaluate and contrast these two methods in a realistic, low-resourced setup, where the budget for annotation is extremely limited, and second to reduce the training time and assess the practicality of our in-context probing approach. 

As previous experiments suggest, the performance of ICL is highly dependent on the instruction. For the purpose of this experiment, we plot the average performance of different methods on the $5$ instructions. As in previous experiments, we sample training data (or demonstration examples) with $5$ different seeds, and visualize the standard deviation in the performance in \cref{fig:q3}.

We observe that for \tweeteval{}, probing larger models (e.g., \flanxxl{}) with only $40$ training examples performs competitive to ICL. As discussed earlier, ICL is not very effective on smaller models, therefore, with as few as $40$ training examples, ICP outperforms ICL significantly (see \flanbase{} and \flans{} results). The same conclusion holds for other tasks, see \cref{fig:sample-eff-all} for further experiments.

\paragraph{ICP vs. Probing.} We investigate the role of \emph{contextualization} in probe's performance by comparing the sample efficiency of in-context probing to probing. Different from ICP, where we contextualize the input with the instruction, with probing we only probe the input's representation. We compare probing and ICP on \flanxxl{} across the three different tasks in \cref{fig:remove-inst}.\footnote{To reduce the clutter in the figure, we only visualize the maximum F1 score of decoding and calibrated decoding methods.} In \tweeteval{} and \climate{} tasks, both ICP and probing reach the decoding baselines with using only $40$ training samples. Interestingly, in the \mqp{} task, ICP is significantly more sample efficient than probing; it matches the performance of decoding baselines with $40$ training samples, while probing does the same with $80$ samples.   

\subsection{In-Context Probing vs. Finetuning} \label{sec:finetune}
We compare in-context probing with prompt-based finetuning. To ensure a fair comparison, we set the training budget to $200$ examples and provide a head-to-head comparison between finetuning and ICP in \cref{tab:ft}. We use the same set of $5$ instructions to finetune models with prompts. Therefore, for each task and model size, we finetune the model $5$ times, each time with a unique instruction. We measure the average performance and the standard deviation of these $5$ training runs on the test set and compare that to measurements from probing models obtained in the same experimental setup.

We observe that finetuning and ICP are comparably robust to the change in instructions they are trained with. This is reflected in low standard deviation across the $5$ runs. Importantly, ICP is competitive with finetuning across different model sizes and tasks. This is particularly promising, given that the probes have up to $6$ orders of magnitude less trainable parameters compared to their corresponding finetuned models, and therefore are faster to train and need less memory. 

\begin{table}[h]
    \centering
    \adjustbox{width=\columnwidth}{%
    \begin{tabular}{@{}lp{2cm}ccc@{}}\toprule
    & \# parameters & \mqp{} & \tweeteval{} & \climate{} \\ \midrule
    ICP-\att & $67$M & $89.22\stdtiny{0.68}$  & $71.64\stdtiny{1.22}$  & $53.92\stdtiny{2.19}$ \\
    ICP-\reg & $10$K & $\mathbf{89.95}\stdtiny{0.56}$  & $\mathbf{72.95}\stdtiny{\mathbf{0.59}}$  & $\mathbf{59.34}\stdtiny{\mathbf{1.21}}$ \\ \bottomrule
    \end{tabular}}
    \caption{Comparing F1 Macro (\%) of attentional probe (\att) to logistic regression probe (\reg) suggests that adding complexity to the probe does not improve the performance on the classification tasks.}
    \label{tab:att-reg}
\end{table}
\subsection{Logistic Regression vs. Attentional Probe} \label{sec:arch}
In this experiment we investigate the effect of the probes' complexity on the performance. To this end, we compare the performance of the logistic regression ($\reg$) probe with the attentional probe ($\att$) that has more trainable parameters. \cref{tab:att-reg} shows the results on \flanxxl{} model and \cref{tab:ft-complete} includes the results on all \flant{} models. We find that logistic regression probes are as effective as attentional probes, while using less compute.

\section{Discussion}

With the current progress, and growing accessibility, LLMs are increasingly used to solve a wide range of classification tasks. Often in impromptu and low-data-budget scenarios. 
While in-context learning offers an efficient option to deploy LLMs on new tasks, it is unstable: Our results confirm that instructions formulation can significantly affect results. We propose in-context probing as a systematic approach to reduce instability. Our probing method is extremely simple and only requires access to the output representations of LLM's last layer.

Real-world classification tasks often rely only on a few hundred annotated examples between training and validation. Our experiments suggest that even in such settings, in-context probing produces better classifiers on top of smaller models. Furthermore, probing achieves competitive results to ICL in larger models, while being significantly more robust to variations in the instructions.

We hypothesize that probing may be useful also in the development of some of the larger models, to estimate headroom and sensitivity to instruction phrasings in classification tasks. We leave this for future work.

\section{Conclusion}
We propose in-context probing as an alternative to in-context learning. We train light-weight classifiers on top of contextualized representations extracted from \flant{} encoders on $3$ sentence classification tasks. Our experiments suggest that in-context probing is significantly more stable to variations of instructions compared to ICL and calibrated ICL.
With as few as a hundred examples, probing achieves competitive results to ICL on large models, while outperforming ICL on smaller ones.


\section*{Limitations}
\paragraph{Decoder-Only Models.} All experiments are done on $\flant{}$ family of encoder-decoder models. In-context probing can be applied to decoder-only models by probing the representations extracted from the last decoder layer. Further experiments are needed to analyse the effectiveness of in-context probing of decoder-only models.

\paragraph{Probing Larger Language Models.} It is not obvious how the results in this paper transfer to models with trillions of parameters. Since access to such models are limited through APIs, we do not provide probing experiments on larger models.

\section*{Ethical Considerations}
We do not believe the approach presented here further amplifies biases that are already present in the datasets and large language models. Therefore, we foresee no ethical concerns in this work.

\section*{Acknowledgements}
We thank Sascha Rothe, Jannis Bulian, Christian Buck, and Nicola De Cao for technical support and fruitful discussions throughout the project. We further thank Fernando Pereira, Wojciech Gajewski, Michelle Chen Huebscher and Lierni Sestorain Saralegui for providing feedback on a preliminary draft of the project.

\bibliography{refs}
\clearpage
\appendix
\onecolumn
\section{Task Instructions} \label{sec:appendix}
In \cref{tab:inst}, we document the instruction templates that are used throughout the paper. The ordering of the instructions are arbitrary.
\begin{table*}[h]
    \centering
    \begin{tabular}{@{}lp{6.0in}@{}}\toprule
    Index & Instruction \\ \midrule
    \rowcolor{myinst} \multicolumn{2}{c}{Task: Paraphrase Detection: Medical Question Pairs (\mqp) } \\
    \texttt{i0} & Are question1 and question2 similar to each other? answer with Yes or No. \\
    \texttt{i1} & Write Yes if question1 and question2 are similar to each other, write No otherwise. \\
    \texttt{i2} & Is the intent of question1 and question2 similar to each other? answer with Yes or No? \\
    \texttt{i3} & Write Yes if the intent of question1 and question2 are similar to each other, write No otherwise. \\
    \texttt{i4} & Is question1 and question2 semantically equivalent? answer with Yes or No. \\ 
    \rowcolor{myinst} \multicolumn{2}{c}{Task: Natural Language Inference: Climate Fever (\climate)} \\
    \texttt{i0} & Based on the evidence, can we conclude that the claim is definitely supported? Answer with only one of the following options: yes, supported | no, refuted | not enough info \\
    \texttt{i1} & Based on the evidence, can we conclude that the claim is definitely supported? Answer with only one of the following options: yes, supported | no, refuted | no, not enough info \\
    \texttt{i2} & Based on the evidence, can we conclude that the claim is supported? Options: yes, supported | no, refuted | not enough info \\
    \texttt{i3} & Does the evidence support the claim, refute it, or there is no enough info to judge this? Answer with only one of the following options: supports | refutes | not enough info \\
    \texttt{i4} & Given the evidence, is the claim definitely supported, refuted, or there is not enough info? Answer with only one of the following options: supports | refutes | not enough info \\ 
    \rowcolor{myinst} \multicolumn{2}{c}{Task: Hate Speech Detection (\tweeteval)} \\
    \texttt{i0} & Write Yes if you detect something hateful in the tweet write No otherwise. \\
    \texttt{i1} & detect something hateful in the tweet? Yes or No? \\
    \texttt{i2} & Say Yes if the following tweet contains hate speech, say No otherwise. \\
    \texttt{i3} & is there something bad in the tweet? Yes or No? \\
    \texttt{i4} & does the following tweet contain hate speech? Yes or No? \\ \bottomrule
    \end{tabular}
    \caption{Instruction used for each task. All instructions aim to describe the task. However, one can see different levels of expressiveness used in the instructions.}
    \label{tab:inst}
\end{table*}
\section{Dataset Statistics}
We follow the choice of \citet{min-etal-2022-rethinking} and select $3$ diverse datasets. We use the Huggingface version of the data \cite{lhoest-etal-2021-datasets}. We use these datasets to train and evaluate the performance of models on classification tasks, which is in line with the intended use of the data. See \cref{tab:dataset} for dataset statistics. 
\section{Experimental Details} \label{sec:details}
\paragraph{Logistic Regression Probe.} We use scikit-learn \cite{sklearn} implementation of logistic regression. The tolerance for stopping criteria is set to $1e^{-4}$ . We further adjust the class weights by their frequency in the training data, i.e., we set the weight parameter to ``balanced'' in scikit-learn. 
\paragraph{Attentional Probe.} We implement a one layer attention with one attention head. We apply a grid search to find the best learning rate, which is set to $5e^{-6}$ for \flanxxl{}, \flanxl{}, and \flanl{} models, and $5e^{-5}$ for \flanbase{} and \flans{} models. We use Adam optimizer \cite{adam}.
\begin{table}[h]
    \centering
    \begin{tabular}{@{}lcc@{}}\toprule
        Dataset & Training-Set Size & Evaluation-Set Size \\ \midrule
        medical\_questions\_pairs (\mqp{}) & $2438$ & $610$ \\
        climate\_fever (\climate{}) & 1228 & 307 \\
        tweet\_eval-hate (\tweeteval{}) & 8993 & 999 \\ \bottomrule
    \end{tabular}
    \caption{Dataset Statistics}
    \label{tab:dataset}
\end{table}
\section{Experimental Results} \label{sec:experiments-extra}
\begin{figure}
     \centering
     \begin{subfigure}[b]{\textwidth}
         \centering
         \includegraphics[width=0.7\textwidth]{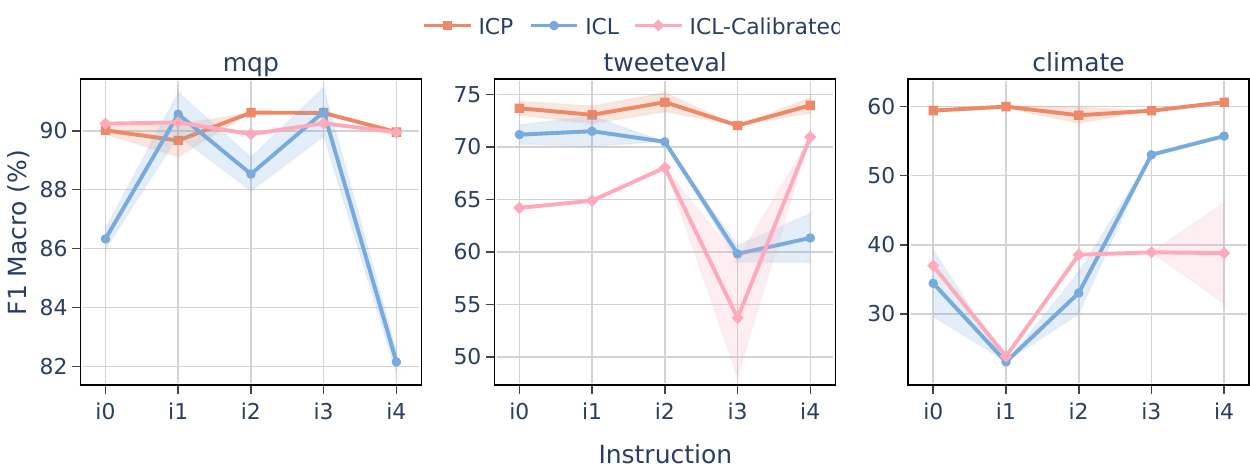}
         \caption{\flanxxl{}}
     \end{subfigure}
     \hfill
     \begin{subfigure}[b]{\textwidth}
         \centering
         \includegraphics[width=0.7\textwidth]{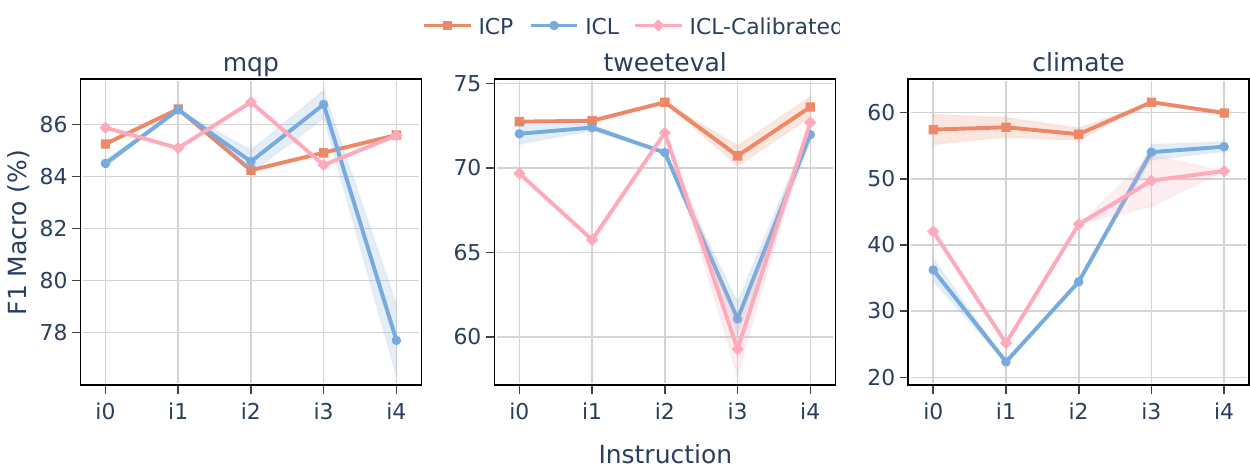}
         \caption{\flanxl{}}
     \end{subfigure}
     \hfill
     \begin{subfigure}[b]{\textwidth}
         \centering
         \includegraphics[width=0.7\textwidth]{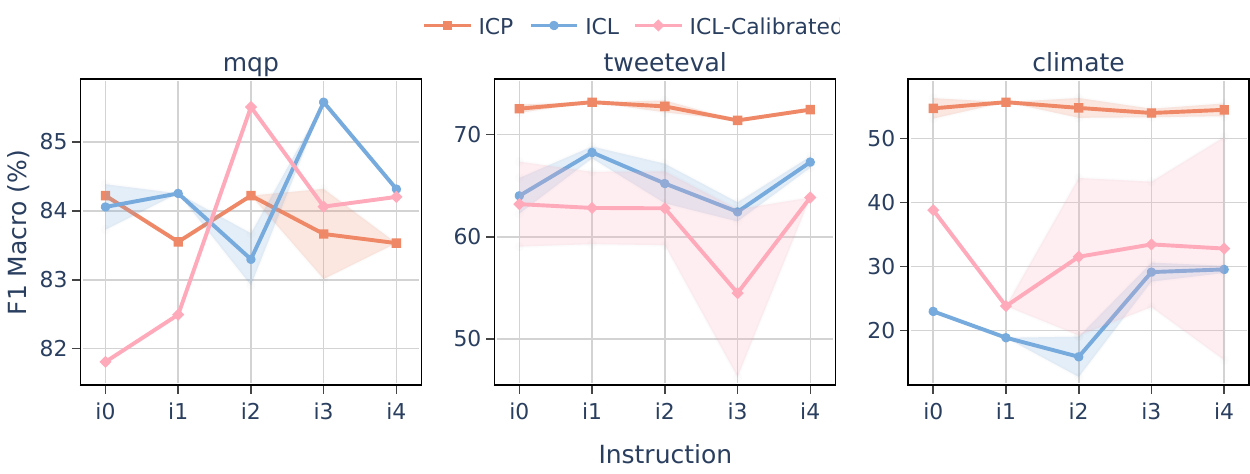}
         \caption{\flanl{}}
     \end{subfigure}
     \begin{subfigure}[b]{\textwidth}
         \centering
         \includegraphics[width=0.7\textwidth]{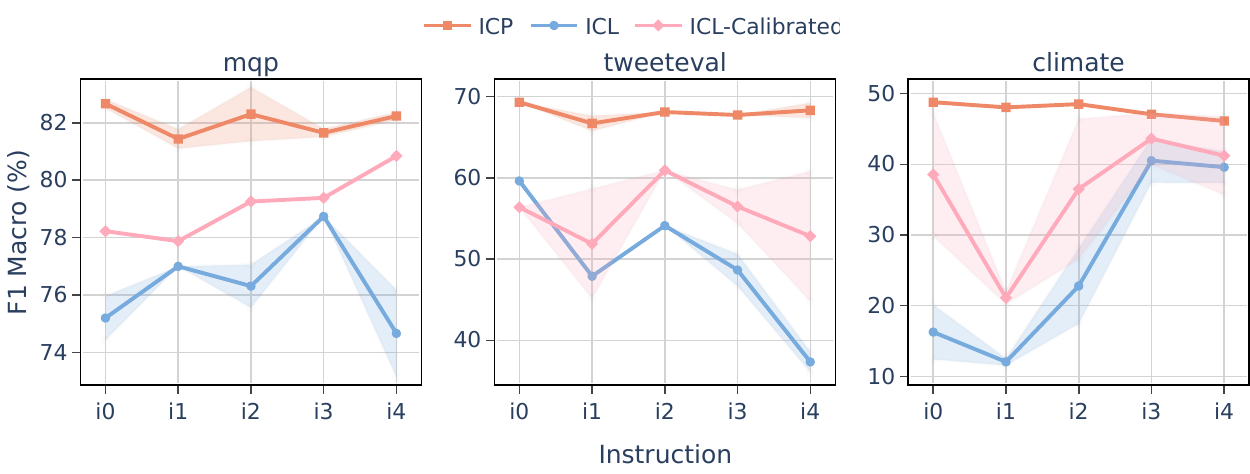}
         \caption{\flanbase{}}
     \end{subfigure}
     \begin{subfigure}[b]{\textwidth}
         \centering
         \includegraphics[width=0.7\textwidth]{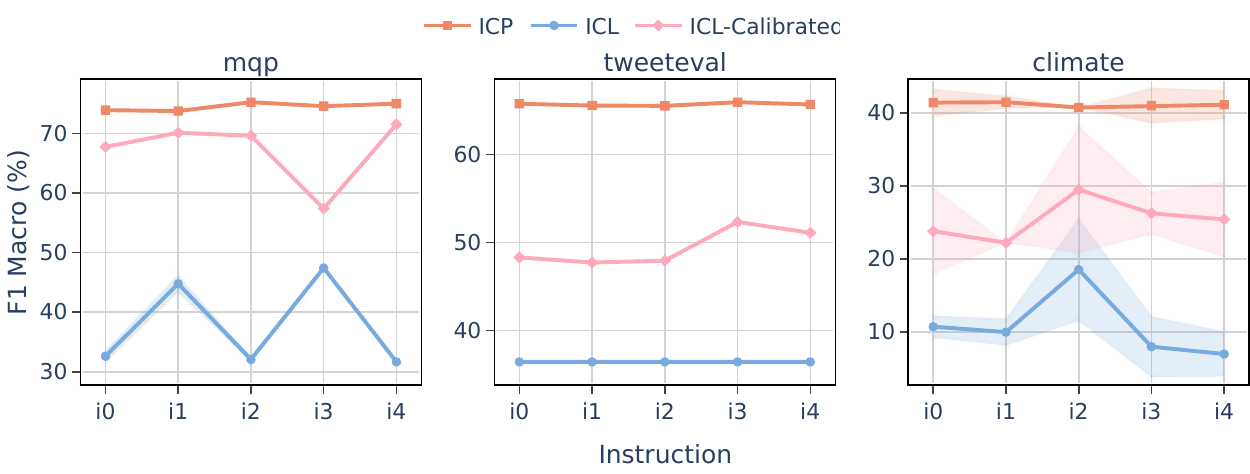}
         \caption{\flans{}}
     \end{subfigure}
        \caption{Comparing the robustness of different approaches to the change of instructions across models and tasks. Across all of the tasks and model sizes, in-context probing offers competitive performance to ICL and is more robust to instruction change.}
        \label{fig:robust-all}
\end{figure}
\begin{figure}
     \centering
     \begin{subfigure}[b]{\textwidth}
         \centering
         \includegraphics[width=0.8\textwidth]{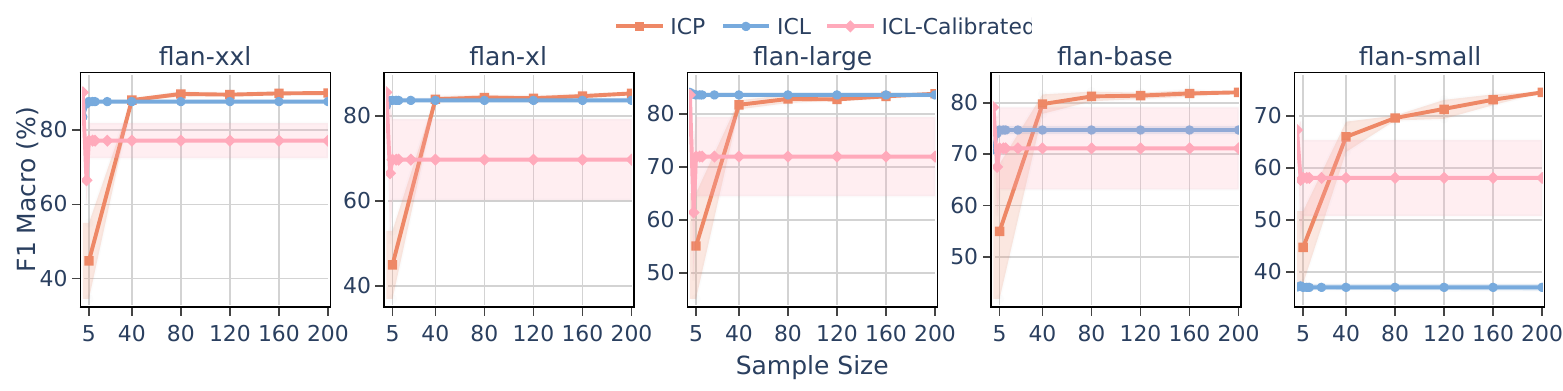}
         \caption{\mqp{}}
     \end{subfigure}
     \hfill
     \begin{subfigure}[b]{\textwidth}
         \centering
         \includegraphics[width=0.8\textwidth]{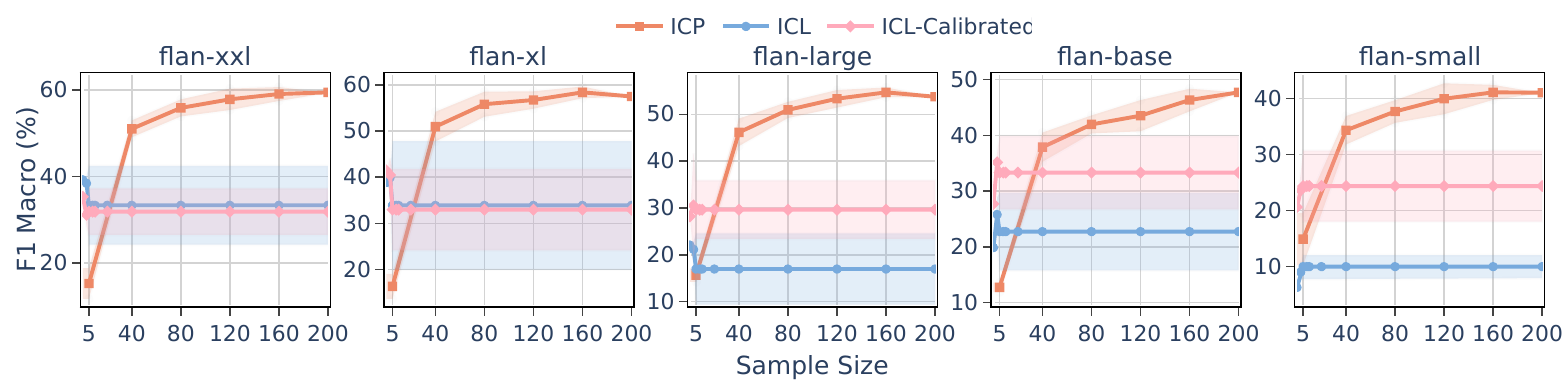}
         \caption{\climate{}}
     \end{subfigure}
     \hfill
     \begin{subfigure}[b]{\textwidth}
         \centering
         \includegraphics[width=0.8\textwidth]{figs/figures-raw/samples-tweeteval_svg-raw.pdf}
         \caption{\tweeteval{}}
     \end{subfigure}
        \caption{Measuring samples efficiency of ICP in different tasks and models. In all models and tasks, ICP matches the performance of ICL and calibrated ICL after only training on a hundred training examples.}
        \label{fig:sample-eff-all}
\end{figure}

\begin{table*}[t]
    \centering
    \adjustbox{width=\textwidth}{%
    \begin{tabular}{@{}lccccccc@{}}\toprule
    & \multirow{2}{2cm}{\# trainable parameters}& \multicolumn{2}{c}{\mqp{}} & \multicolumn{2}{c}{\tweeteval{}} & \multicolumn{2}{c}{\climate{}} \\ \cmidrule(l){3-4} \cmidrule(l){5-6} \cmidrule(l){7-8}
    & & F1 Macro ($\%$) $\uparrow$ & std. $\downarrow$  & F1 Macro ($\%$) $\uparrow$ & std. $\downarrow$ & F1 Macro ($\%$) $\uparrow$ & std. $\downarrow$ \\ 
    \rowcolor{myinst} \multicolumn{8}{c}{\flanxxl{}} \\
    FT & $11$B & $89.12$ & $\mathbf{0.53}$  & $71.0$ & $3.14$ & $54.60$ & $3.67$ \\
    ICP-\att& $67$M & $89.22$ & $0.68$ & $71.64$ & $1.22$ & $53.92$ & $2.19$\\ 
    ICP-\reg& $10$K & $\mathbf{89.95}$ & $0.56$ & $\mathbf{72.95}$ & $\mathbf{0.59}$ & $\mathbf{59.34}$ & $\mathbf{1.21}$\\ 
    \rowcolor{myinst} \multicolumn{8}{c}{\flanxl{}} \\
    FT & $3$B & $\mathbf{87.89}$ & $\mathbf{0.38}$ &  $69.47$ & $2.52$ & $50.91$ & $\mathbf{1.49}$ \\
    ICP-\att& $16$M & $85.66$ & $0.87$ & $71.57$ & $1.26$ & $52.13$ & $2.44$\\ 
    ICP-\reg& $6$K & $85.32$ & $0.78$ & $\mathbf{72.66}$ & $\mathbf{1.1}$ & $\mathbf{57.45}$ & $2.72$\\ 
    \rowcolor{myinst} \multicolumn{8}{c}{\flanl{}} \\
    FT & $780$M & $\mathbf{86.23}$ & $0.85$  & $69.26$ & $1.2$  & $51.37$ & $5.1$\\
    ICP-\att& $4$M & $84.17$ & $0.52$ & $70.06$ & $\mathbf{0.5}$ & $29.80$ & $3.48$\\ 
    ICP-\reg& $3$K & $83.81$ & $\mathbf{0.33}$ & $\mathbf{72.42}$ & $0.58$ & $\mathbf{53.75}$ & $\mathbf{1.50}$\\ 
    \rowcolor{myinst} \multicolumn{8}{c}{\flanbase{}} \\
    FT & $250$M & $79.57$ & $1.72$  & $64.36$ & $\mathbf{0.49}$  & $45.58$ & $5.17$ \\
    ICP-\att& $2$M & $79.06$ & $1.65$ & $66.54$ & $1.73$ & $41.53$ & $2.17$\\ 
    ICP-\reg& $2$K & $\mathbf{82.08}$ & $\mathbf{0.59}$ & $\mathbf{67.92}$ & $0.93$ & $\mathbf{47.71}$ & $\mathbf{0.98}$\\ 
    \rowcolor{myinst} \multicolumn{8}{c}{\flans{}} \\
    FT & $80$M & $71.73$ & $0.80$  & $55.35$ & $4.87$  & $34.98$ & $1.63$ \\
    ICP-\att& $1$M & $59.80$ & $1.56$ & $63.08$ & $2.63$ & $38.03$ & $1.42$\\ 
    ICP-\reg& $1$K & $\mathbf{74.44}$ & $\mathbf{0.77}$ & $\mathbf{65.73}$ & $\mathbf{0.17}$ & $\mathbf{40.92}$ & $\mathbf{0.47}$\\ \bottomrule
    \end{tabular}}
    \caption{Comparing the performance of in-context probing (ICP) to finetuning (FT) with $200$ training examples. The standard deviation (std) shows the performance change with respect to change of instructions used for finetuning or training the probe. While probes have significantly less trainable parameters, they perform competitive (or superior) to finetuning, with comparable standard deviation across different instructions.}
    \label{tab:ft-complete}
\end{table*}

\end{document}